\documentclass[pmlr]{jmlr}

\usepackage{longtable}

\usepackage{booktabs}
\usepackage[load-configurations=version-1]{siunitx} 

\theorembodyfont{\upshape}
\theoremheaderfont{\scshape}
\theorempostheader{:}
\theoremsep{\newline}

\jmlrvolume{NeurIPS 2021 Competition and Demonstration Track}
\jmlryear{2021}
\jmlrworkshop{PMLR}

\usepackage{times}
\usepackage{latexsym}
\usepackage{url}
\usepackage{booktabs}       
\usepackage{amsfonts}       
\usepackage{nicefrac}       
\usepackage{microtype}      

\usepackage{algorithmic}
\usepackage{algorithm}
\usepackage{wrapfig}
\usepackage{lipsum}
\usepackage{xspace}
\usepackage{amsmath}
\usepackage{bm}
\usepackage{graphicx} 
\usepackage[skip=5pt]{caption}
\usepackage{paralist}
\usepackage{float}
\usepackage{multirow}
\usepackage{diagbox}
\usepackage{tikz}
\usepackage{booktabs}
\usepackage{array, makecell} 
\usepackage{natbib}
\usepackage{amsmath,amssymb}
\usepackage{enumitem}
\usepackage{acronym}
\usepackage{wrapfig}

\newcounter{todocnt}

\newcommand{\name}{IGLU\xspace}

\acrodef{RL}{Reinforcement Learning}
\acrodef{RNN}{Recurrent Neural Net}

\title{Interactive Grounded Language Understanding \\ in a Collaborative Environment: IGLU 2021}

\author{\Name{Julia Kiseleva}\textsuperscript{1}       
        \Name{Ziming Li}\textsuperscript{3}            
        \Name{Mohammad Aliannejadi}\textsuperscript{2} 
        \Name{Shrestha Mohanty}\textsuperscript{1}     
        \Name{Maartje ter Hoeve}\textsuperscript{2}    
        \Name{Mikhail Burtsev}\textsuperscript{4,5}    
        \Name{Alexey Skrynnik}\textsuperscript{5}      
        \Name{Artem Zholus}\textsuperscript{4}         
        \Name{Aleksandr Panov}\textsuperscript{4,5}      
        \Name{Kavya Srinet}\textsuperscript{6}         
        \Name{Arthur Szlam}\textsuperscript{6}         
        \Name{Yuxuan Sun}\textsuperscript{6}           
        \Name{Marc-Alexandre Côté}\textsuperscript{1}
        \Name{Katja Hofmann}\textsuperscript{1}        
        \Name{Ahmed Awadallah}\textsuperscript{1} \\
        \\
        \Name{Linar Abdrazakov}\textsuperscript{7}     
        \Name{Igor Churin}\textsuperscript{8}          
        \Name{Putra Manggala}\textsuperscript{2}       
        \Name{Kata Naszadi}\textsuperscript{2}         
        \Name{Michiel van der Meer}\textsuperscript{10}
        \Name{Taewoon Kim}\textsuperscript{11} \\
        \\
\Email{\tt julia.kiseleva@microsoft.com}\\
\\
\addr
\textsuperscript{1}Microsoft Research \\ 
\textsuperscript{2}University of Amsterdam \\
\textsuperscript{3}Alexa AI \\
\textsuperscript{4}MIPT \\
\textsuperscript{5}AIRI  \\
\textsuperscript{6}Facebook AI \\ 
\textsuperscript{7}Moscow Institute of Physics and Technology \\
\textsuperscript{8}Kalashnikov Izhevsk State Technical University \\
\textsuperscript{10}Universiteit Leiden \\
\textsuperscript{11}Vrije Universiteit Amsterdam
}

\editors{Douwe Kiela, Marco Ciccone, Barbara Caputo}

\begin{document}
\maketitle

\begin{abstract}

Human intelligence has the remarkable ability to quickly adapt to new tasks and environments. Starting from a very young age, humans acquire new skills and learn how to solve new tasks either by imitating the behavior of others or by following provided natural language instructions. To facilitate research in this direction, we propose \emph{IGLU: Interactive Grounded Language Understanding in a Collaborative Environment}.
The primary goal of the competition is to approach the problem of how to build interactive agents that learn to solve a task while provided with grounded natural language instructions in a collaborative environment. Understanding the complexity of the challenge, we split it into sub-tasks to make it feasible for participants.  

\end{abstract}

\subsection*{Keywords}
Natural Language Understanding (NLU), Reinforcement Learning (RL), Grounded Learning, Interactive Learning, Games

\section{Introduction}

Humans have the remarkable ability to quickly adapt to new tasks and environments. Starting from a very young age, humans acquire new skills and learn how to solve new tasks either by imitating behavior of others or by following natural language instructions that are provided to them~\citep{an_imitation_1988, council_how_1999}. 
Natural language communication provides a natural way for humans to acquire new knowledge, enabling us to learn quickly through language instructions and other forms of interaction such as visual cues. This form of learning can even accelerate the acquisition of new skills by avoiding trial-and-error and statistical generalization when learning only from observations~\citep{thomaz2019interaction}. Studies in developmental psychology have shown evidence of human communication being an effective method for transmission of generic knowledge between individuals as young as infants~\citep{csibra2009natural}. These observations have inspired attempts from the AI research community to develop grounded interactive \emph{agents} that are capable of engaging in natural back-and-forth dialog with humans to assist them in completing real-world tasks~\citep{winograd1971procedures,narayan2017towards, levinson2019tom,chen2020ask}.

Importantly, the agent needs to understand when to initiate feedback requests if communication fails or instructions are not clear and requires learning new domain-specific vocabulary~\citep{Aliannejadi_convAI3,rao2018learning, narayan2019collaborative, jayannavar-etal-2020-learning}.
Despite all these efforts, the task is far from solved. 
For that reason, we propose the \name competition, which stands for \emph{Interactive Grounded Language Understanding in a collaborative environment}.

Specifically, the goal of our competition is to approach the following scientific challenge: 
\emph{How to build interactive agents that learn to solve a task while provided with grounded natural language instructions in a collaborative environment?}

By \textit{`interactive agent'} we mean that the agent is able to follow the instructions correctly, is able to ask for clarification when needed, and is able to quickly adapt newly acquired skills, just like humans are able to do while collaboratively interacting with each other.

The described research challenge is naturally related, but not limited, to two fields of study that are highly relevant to the NeurIPS community: Natural Language Understanding and Generation (NLU / NLG) and Reinforcement Learning (RL).
\section{Background and Related Work}

\paragraph{Relevance of NLU/G} 
Natural language interfaces (NLIs) have been the ``holy grail'' of human-computer interaction and information search for decades~\citep{woods1972lunar, codd1974seven, hendrix1978developing}. The recent advances in language understanding capability~\citep{devlin2018bert, LiuRoberta_2019, clark2020electra, adiwardana2020towards, roller2020recipes, brown2020language}  powered by large-scale deep learning and increasing demand for new applications has led to a major resurgence of natural language interfaces in the form of virtual assistants, dialog systems,  semantic  parsing, and question answering systems~\citep{liu2017iterative, liu2018adversarial, dinan2020second, zhang2019dialogpt}. The horizon of NLIs has also been significantly  expanding  from databases~\citep{copestake1990natural} to,  knowledge  bases~\citep{berant2013semantic}, robots~\citep{tellex2011understanding}, Internet of Things (virtual  assistants like  Siri  and Alexa), Web service APIs~\citep{su2017building}, and other forms of interaction~\citep{fast2018iris, desai2016program, young2013pomdp}.
Recent efforts have also focused on interactivity and continuous learning to enable agents to interact with users to resolve the knowledge gap between them for better accuracy and transparency. This includes systems that can learn new task from instructions~\citep{li-etal-2020-interactive}, assess their uncertainty~\citep{yao-etal-2019-model}, ask clarifying questions~\citep{ Aliannejadi_convAI3, aliannejadi2021building} and seek and leverage feedback from humans to correct mistakes ~\citep{elgohary-etal-2020-speak}.

\paragraph{Relevance of RL} Recently developed RL methods celebrated successes for a number of tasks~\citep{bellemare2013arcade, mnih2015human, mnih2016asynchronous, silver2017mastering, hessel2018rainbow}. One of the aspects that helped to speed up RL methods development is game-based environments, which provide clear goals for an agent to achieve in flexible training settings. However, training RL agents that can follow human instructions has  attracted fewer exploration~\citep{chevalier2019babyai,cideron2019self,hu2019hierarchical, chen2020ask, shu2017hierarchical}, due to complexity of the task and lack of proper experimental environments.
~\citet{shu2017hierarchical} proposed a hierarchical policy modulated by a stochastic temporal grammar for efficient multi-task reinforcement learning where each learned task corresponds to a human language description in Minecraft environment. The BabyAI platform~\citep{chevalier2019babyai} aims to support investigations towards learning to perform language instructions with a simulated human in the loop. ~\citet{chen2020ask} demonstrated that using step-by-step human demonstrations in the form of natural language instructions and action trajectories can facilitate the decomposition of complex tasks in a crafting environment.

\paragraph{Minecraft as an Environment for Grounded Language Understanding}
\citet{szlam_why_2019} substantiated the advantages of building an open interactive assistant in the sandbox construction game of Minecraft instead of a ``real world'' assistant, which is inherently complex and inherently costly to develop and maintain. The Minecraft world's constraints (e.g., coarse 3-d voxel grid and simple physics) and the regularities in the head of the distribution of in-game tasks allow numerous scenarios for grounded NLU research~\citep{yao2020imitation, srinet-etal-2020-craftassist}. Minecraft is an appealing competition domain due to its popularity as a video game, of all games ever released, it has the second-most total copies sold. Moreover, since it is a popular game environment, we can expect players to enjoy interacting with the assistants as they are developed, yielding a rich resource for a human-in-the-loop studies.
Another important advantage of using Minecraft is the availability of the highly developed set of tools for logging agents interactions and deploying agents for evaluation with human-in-the-loop, including: 
\begin{itemize}[nosep, leftmargin=*]
    \item \textit{Malmo}~\citep{johnson2016malmo}: a powerful platform for AI experimentation;
    \item \textit{Craftassist}~\citep{gray_craftassist_2019}: a framework for dialog-enabled interactive agents\footnote{CraftAssist is utilized for data collection};
    \item \textit{TaskWorldMod}~\citep{ogawa-etal-2020-gamification}: a platform for situated task-oriented dialog data collection using gamification; and
    \item \textit{MC-Saar-Instruct}~\citep{kohn2020mc}: a platform for Minecraft Instruction Giving Agents.
\end{itemize}

\noindent
Besides, mainly due to the success of previous competitions~\citep{guss2019minerlcomp,perez2019multi}, Minecraft is a widely used environment by the RL community for experimentation with (mainly single) agents trained by demonstration. Therefore, using Minecraft would set a low barrier for the RL community to contribute to \name. To simplify the competition settings and possibly lower the entry bar for the NLU/NLG community, we will use simulated Blocks World in Minecraft~\citep{jayannavar-etal-2020-learning}.

\paragraph{Relevance to Real Live Scenarios and Societal Impact}
Several important real-life scenarios have the potential to benefit from the results of our competition:
\begin{itemize}[leftmargin=*]
    \item \textbf{Education:} 
    \textit{Minecraft: Education Edition}\footnote{\url{https://education.minecraft.net/}} is a game-based learning platform that promotes creativity, collaboration, and problem-solving in an immersive digital environment. As of 2021, educators in more than $115$ countries are using Minecraft across the curriculum. As stated in~\citet{url-minecraft-edu}, adding AI elements to this educational platform will move its potential to a new level. AI applications have the power to become a great equalizer in education. Students can get personalized education and scaffolding while being less dependent on uncontrollable factors such as the quality of their teachers or the amount of help they receive from their caregivers.

    \item \textbf{Robotics:} \citet{bisk-etal-2016-natural} proposed a protocol and framework for collecting data on human-robot interaction through natural language. The work demonstrated the potential for unrestricted contextually grounded communications between human and robots in blocks world. Developing robots to assist humans in different tasks at home has attracted much attention in the Robotics field~\citep{stuckler2012robocup}. In fact, the Robocup@Home\footnote{\url{https://athome.robocup.org/}} and the Room-Across-Room\footnote{\url{https://ai.google.com/research/rxr/habitat}} have run for several years.
    Given that the main human-robot interaction is through dialog, and the robot is supposed to assist the human in multiple tasks, we envision \name to enable more effective task grounded dialog training between human and robots.
\end{itemize}

There is a long history of competitions focused on NLU/G tasks. Especially in recent years we have seen a large number of challenges dedicated to open-domain dialog systems~\citep{10.1145/3465272,scai-2020-international,spina2019cair,chuklin2018proceedings,arguello2018second}, such as ConvAI~\citep{burtsev2020conversational}, ConvAI2~\citep{dinan2020second}, ConvAI3: Clarifying Questions for Open-Domain Dialogue Systems (ClariQ)~\citep{Aliannejadi_convAI3,aliannejadi2021building}, as well as a series of competitions of the Alexa Prize\footnote{\url{https://developer.amazon.com/alexaprize}}. There are great efforts in the community to advance task-oriented dialogs by suggesting competitions, such as the Dialog System Technology Challenge (DSTC-8)~\citep{kim2019eighth}; benchmarks and experimental platforms, e.g., Convlab, which offers the annotated MultiWOZ dataset~\citep{budzianowski2018multiwoz} and associated pre-trained reference models~\citep{lee2019convlab}. There are fewer attempts to study multi-modal dialog systems, e.g., Situated Interactive Multi-Modal Conversational Data Collection And Evaluation Platform (SIMMC)~\citep{crook2019simmc} or Audio Visual Scene-Aware Dialog Track~\citep{hori2018audio}.

There are a number of RL competitions such as MineRL~\citep{guss2019minerlcomp} and MARLO~\citep{perez2019multi} that leverage the Minecraft environment. RL approaches have also been tried for text games environments, such as TextWold~\citep{yuan2019interactive}\footnote{\url{https://www.microsoft.com/en-us/research/project/textworld/}} and Learning in Interactive Games with Humans and Text(Light)~\citep{urbanek2019learning}\footnote{\url{https://parl.ai/projects/light/}}.

In comparison with previous efforts, to our knowledge, we are the first to propose a competition that tackles the task of grounded language understanding and interactive learning that brings together the NLU/G and RL research communities. The other key difference is our attempt to perform a human-in-the-loop evaluation as a final way for evaluating. 
\section{Data}

\paragraph{The general setup} \name is partially motivated by the HCRC Map Task Corpus~\citep{thompson_hcrc_1993}, which consists of route-following dialogs between an \emph{Instruction Giver} and a \emph{Follower}. 
\citet{narayan2019collaborative} collected an openly available Minecraft dialog Corpus for a Collaborative Building Task. The authors used the following setup: the Architect is provided with a target structure that needs to be built by the Builder. The Architect provides instructions to the Builder on how to create the target structure and the Builder can ask clarifying questions to the Architect if an instruction is unclear~\cite{zhang-etal-2021-learning}. This dialog happens by means of a chat interface. The Architect is invisible to the Builder, but the Architect can see the actions of the Builder. 
\begin{itemize} [nosep,leftmargin=*]
    \item $509$ collected human-to-human dialogs along with RGB observations, and inventory information;
    \item games played over the course of 3 weeks (approx. 62 hours overall) with $40$ volunteers. Each game took 8.55 minutes on average; and
    \item $163$ tasks for the Architect-Builder collaborative game.
\end{itemize}
This dataset is used at the warm-up stage.
To enable data collection outside of lab settings and bring it to the crowdsourcing platform, we adapted and extended the tool described in~\cite{johnson2016malmo}\footnote{Here is the visualization of the resulted data \url{https://youtu.be/Ls6Wv7EUDA0}}
We record the progression of each task, corresponding to the construction of a target structure by an Architect and Builder pair, as a discrete sequence of game observations. Each observation contains the following information: 1) a time stamp, 2) the chat history up until that point in time, 3) the Builder's position (a tuple of real-valued $x$, $y$, $z$ coordinates as well as pitch and yaw angles, representing the orientation of their camera), 4) the Builder's block inventory, 5) the locations of the blocks in the build region.
The final dataset contains $47$ completed games, where the median duration of the game is $59$ minutes. The collected dataset has $871$ utterances, where the average length of utterance is $19,32$ words. Out of all utterances, $126$ was clarifying questions.

During the competition, the submissions were evaluated in CodaLab. The CodaLab platform allows the submission system's flexible design, leaderboard, metrics, and competition phases. Participants submit the code of their agents optionally with trained model weights and all other necessary files as one zip file. During the evaluation, the automated platform receives a solution, runs several episodes of the IGLU environment or feeds the model with several dataset utterances depending on a track of the competition, and calculates automatic evaluation scores. 
\section{IGLU Tasks}

Given the current state of the field, our main research challenge (i.e., \emph{how to build interactive agents that learn to solve a task while provided with grounded natural language instructions in a collaborative environment}) might be too complex to suggest a reasonable end-to-end solution. Therefore, we split the problem into the following concrete research questions, which correspond to separate tasks that can be used to study each component individually before joining all of them into one system~\citep{jones1988look}:

\begin{enumerate}[label=\textbf{RQ\arabic*},nosep]
 \item \emph{How to teach}? \\ In other words, what is the best strategy for an Architect when instructing a Builder agent, such that the concept is reasonably explained? (The suggested task is presented in Section~\ref{sec:task1}).
 \item \emph{How to learn?} \\
  That is, what methods should be used to train a Builder that can follow given instructions from an Architect? \\
  This question can be further split into two sub-questions:
  \begin{enumerate}[label=\textbf{RQ2.\arabic*},nosep]
   \item \emph{How is a `silent' Builder able to learn?}\\
   A silent Builder follows instructions without the ability to ask for any clarification from the Architect. (The suggested task is presented in Section~\ref{sec:task2}).
   \item \emph{How is an `interactive' Builder able to learn?}\\
   An interactive Builder can ask clarifying questions to the Architect to gain more information about the task in case of uncertainty. (Due to difficulty we leave this task for exploration in future competitions).
  \end{enumerate}
\end{enumerate}

\subsection{Task 1: Architect}
\label{sec:task1}

In this task, our goal is to develop an Architect that can generate appropriate step instructions based on the observations of environment and the Builder's behavior. 
At the beginning of each task, we give all the details of the target structure (e.g., types, colors and coordinated of blocks) to the Architect. The Architect needs to decompose the building process of this compound structure into a sequence of step instructions that the Builder can follow. During the interaction, the Architect has to compare the half-finished structure with the target structure and guide the Builder to complete the building of remaining components via generated instructions. The step instructions can be neither too detailed nor too general. In summary, the Architect is expected to be able to give instructions, correct the Builders’ mistakes and answer their questions by comparing the built structure against the target structure and by understanding the preceding dialog\footnote{We can see the similarity with newly published work on using human instructions to improve generalization in RL~\citep{chen2020ask}.}.

\subsubsection{Task setup}
We aim to generate a suitable Architect utterance, given access to 1) the detailed information of the target structure and 2) the entire game state context leading up to a certain point in a human-human game at which the human Architect spoke next. This task can be seen as a multimodal text generation, where the target structure, the built structure and the dialog history are input and the next Architect's utterance is the output. The model developed for this task can involve both language understanding and visual understanding depending on the methods for world state representations.

\subsubsection{Evaluation and Baseline}
\paragraph{Automatic evaluation}
To evaluate how closely the generated utterances resemble the human utterances, we adopt standard BLEU scores~\citep{papineni2002bleu}. We also make use of the modified \textit{Precision} and \textit{Recall} of domain-specific keywords defined in~\citet{narayan2019collaborative}. The defined keywords are instrumental to task success, including colors, spatial relations, and other words that are highly indicative of dialog actions.

We provide two baselines that are presented in~\cite{narayan2019collaborative}\footnote{The implementation of the baseline can be found using the following repository \url{https://github.com/prashant-jayan21/minecraft-dialogue-models}.}. 
We reproduce the results on the test set presented in~\citep{narayan2019collaborative} in Table~\ref{Table:architect_bleu}. The hyper-parameters of architect models have been fine-tuned on the validation set. By augmenting both the global world state and local world state, Seq2Seq with global and local information managed to show noticeable improvements on each of the automatic metrics. The provided baseline definitely leaves room for improvement. All the architect models will be re-evaluated after we collect a larger dialog corpus.In the course of the competition, none of the teams could outperform the suggested baseline.

 \begin{table}[ht!]
  \centering
  \resizebox{0.7\linewidth}{!}{
  \begin{tabular}{ l c*{9}{c}}
    \toprule
 & \multicolumn{4}{c}{BLEU} & \multicolumn{4}{c}{Precision/ Recall}  \\
  \cmidrule(r){2-5}
  \cmidrule(r){6-9}
 \makecell{Metrics}  &\makecell{BLEU-1} & \makecell{BLEU-2} & \makecell{BLEU-3} & \makecell{BLEU-4} & \makecell{all keywords} &\makecell{colors} &\makecell{spatial}  &\makecell{dialog}\\
\midrule
Seq2Seq    &15.3  &7.8     &4.5    &2.8  &11.8/11.1   &8.1/17.0 &9.3/8.6  &17.9/19.3  \\
\makecell{+global \& local} &15.7  &8.1     &4.8    &2.9  &13.5/14.4   &14.9/28.7  &8.7/8.7  &18.5/19.9  \\
   \bottomrule
  \end{tabular}
  }
\caption{BLEU and term-specific precision and recall scores on the test set, originally reported in \citep{narayan2019collaborative}, which were able to reproduce.} 
\label{Table:architect_bleu}
\vspace{-0.5cm} 
\end{table}

\subsection{Task 2: Silent Builder}
\label{sec:task2}

\begin{wrapfigure}{r}{0.6\textwidth}
   \begin{center}
     \includegraphics[width=0.6\textwidth]{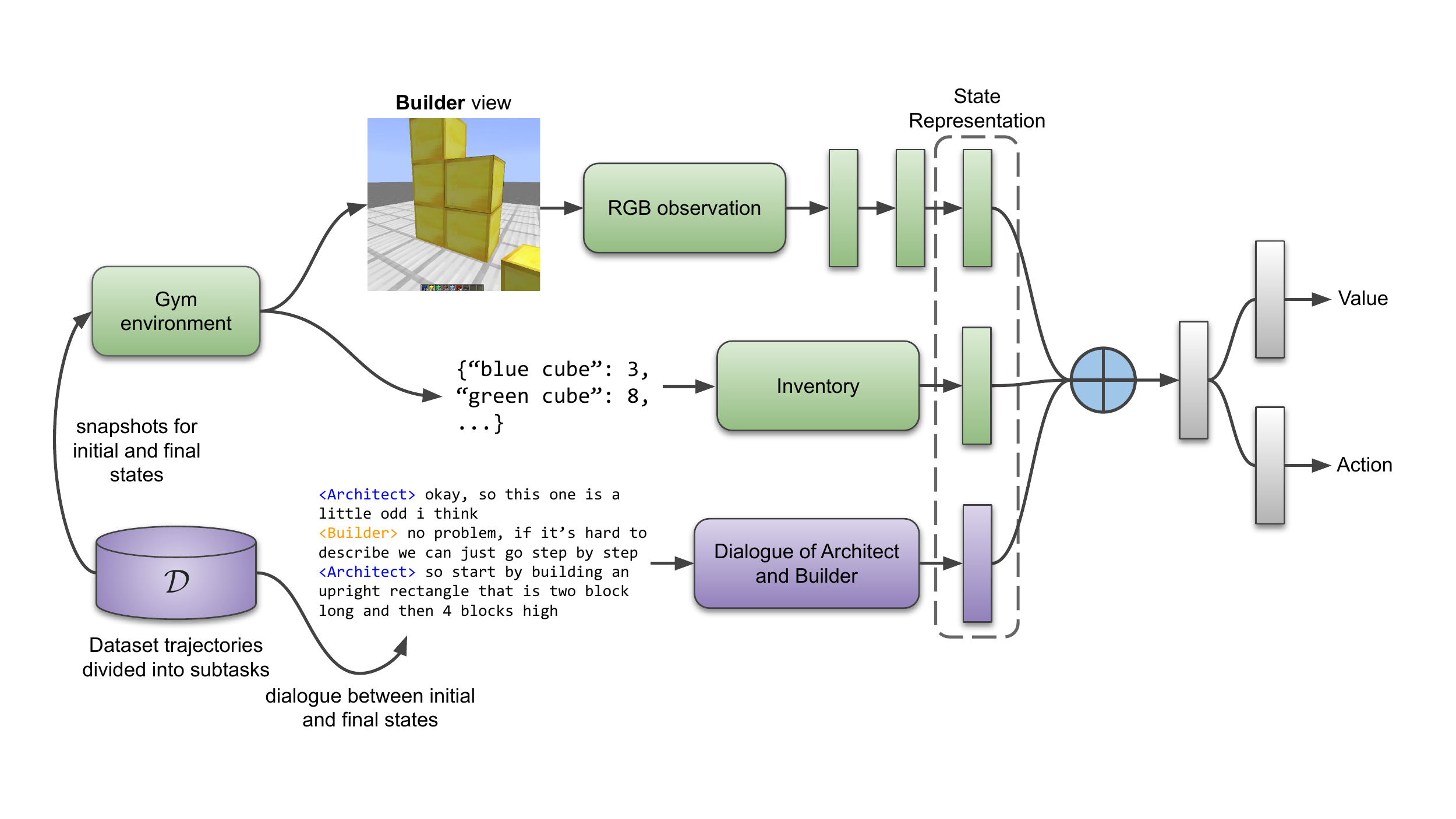}
   \end{center}    
   \caption{The overall pipeline for suggested baselines}
   \label{fig:silent-builder}
\end{wrapfigure}

The overall setup for training initial baselines for the silent Builder that will be used for comparison is presented in Figure~\ref{fig:silent-builder}.

\subsubsection{IGLU Environment}
\label{sec:code_rl}
As the main tool for training grounded agents, we built a gym environment\footnote{\href{https://github.com/iglu-contest/iglu}{https://github.com/iglu-contest/iglu}} that allows training embodied agents in the Minecraft-like gridworld. In the environment, the agent in placed at the center of an empty building zone marked with white blocks. The agent can navigate over the world, place, and break blocks. It has an inventory with six different types of blocks: blue, red, green, orange, purple, and yellow. This setup mimics the data collection setting of \cite{narayan2019collaborative}. The goal of the agent is to analyze the past human-human dialog and reproduce a structure that was a result of this dialog. 

The observation space consists of a POV image $(64, 64, 3)$, inventory item counts $(6,)$, building zone shapshot $(11, 11, 9)$, full dialog as string, and the agent $(X, Y, Z)$ position with two angles (pitch, yaw) $(5,)$. In each cell of the building zone grid component there is an ID that defines the block type in that position (e.g., $0$ for air, $1$ for blue block, etc.). The action space can be chosen from three possible options: human-level actions, distrete actions, and continuous actions. For human-level control, we employ the same action space as is used in MineRL environment \cite{guss2019minerlcomp}. That is, the agent can simulate key press to move forward, backward, left, right, to jump, also it can control the player's camera with a two dimensional continuous vector. It can place a block by simulating right click press and break by simulating a left one. Last, the block type can be chosen using a specific action. For the discrete type of control, we restrict our agent to step over the centers of blocks. The navigation actions are the same but they are discrete. All other actions are the same. The continuous control also changes only navigation actions. With it, the agent can fly freely inside the building zone. The navigation action is a three dimensional continuous vector that describes a shift direction for an agent.

For simplicity of the task, we allow an agent to build a target structure anywhere inside the building zone by comparing all relative translations and rotations of two zones. The environment calculates intersections over two zones and takes the maximum over these intersections. The agent then receives a reward according to the best match if it has been changed (i.e., non-zero reward if and only if the max match was changed since the last action). If the agent gets closer to the target structure in terms of the maximal match, it receives a reward $+2$. If the structure built by the agent moves further away in terms of a maximal match from the target one (e.g., the agent removes a correctly placed block), the agent gets a reward of $-2$. Otherwise, if the agent places/removes a block outside of a partially built structure (i.e., without changing maximal intersections of structures), it receives a reward of $-1$/$+1$ respectively. If the agent moves outside of the building zone, then the environment terminates immediately.

\begin{wrapfigure}{r}{0.6\textwidth}
\floatconts
 {fig:example2}
 {\caption{Example structures for the IGLU environment.}}
 {%
    \subfigure{%
     \label{struct_a}
     \includegraphics[width=0.15\linewidth]{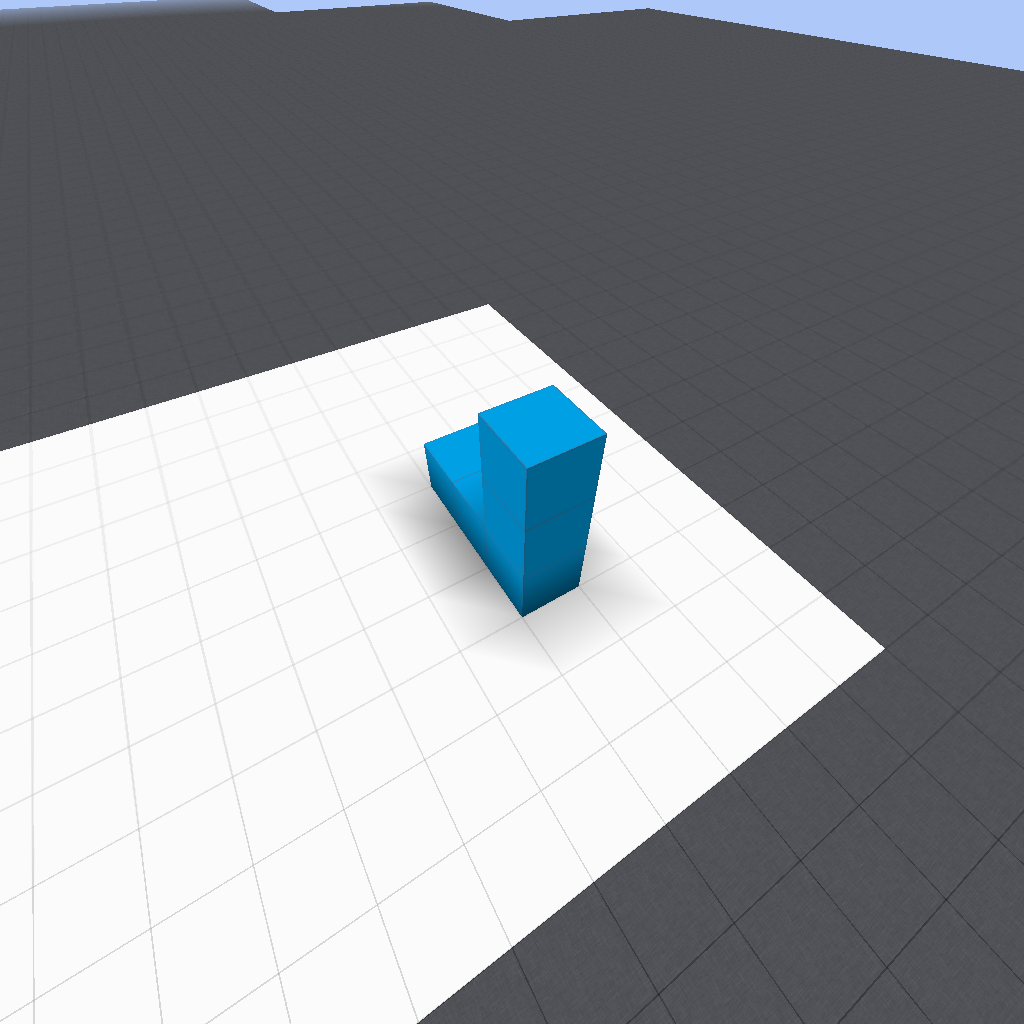}
    }\qquad 
    \subfigure{%
     \label{struct_b}
     \includegraphics[width=0.15\linewidth]{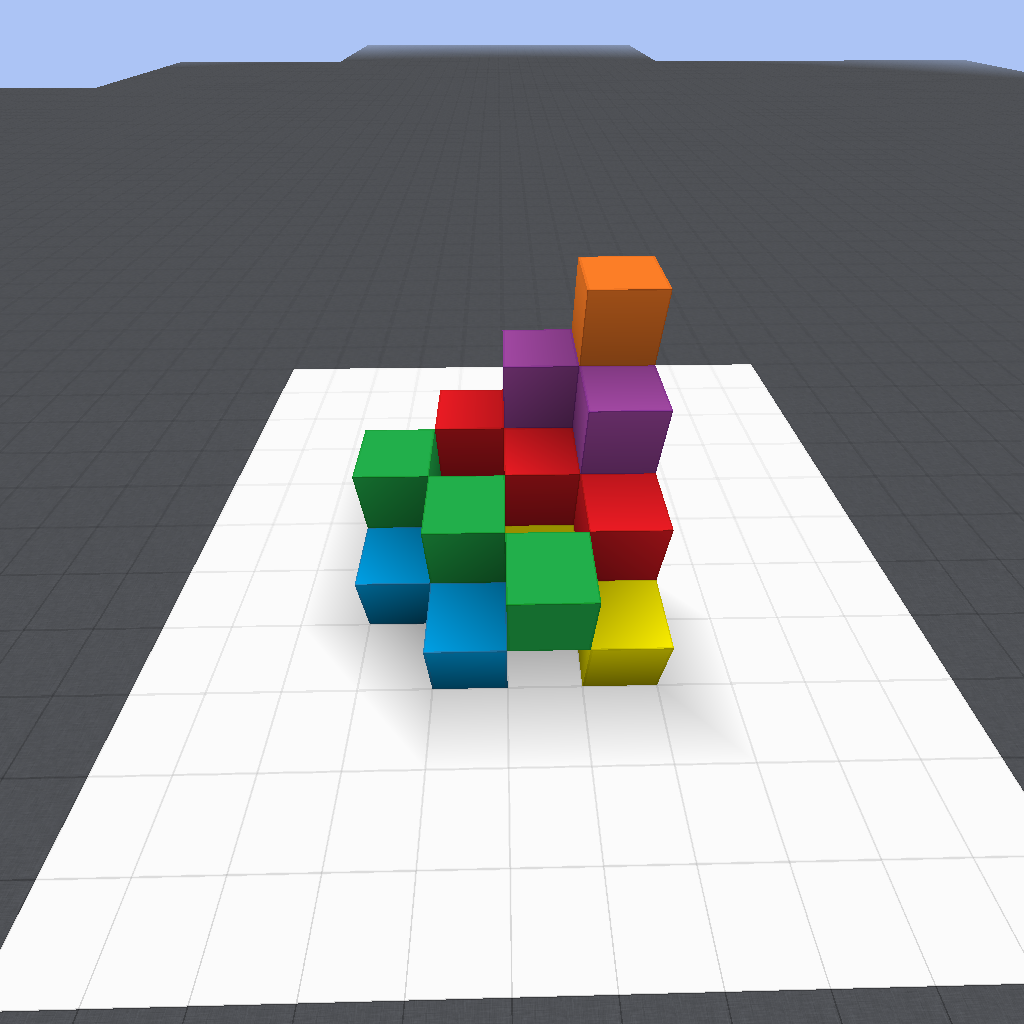}
    }\qquad 
    \subfigure{%
     \label{struct_c}
     \includegraphics[width=0.15\linewidth]{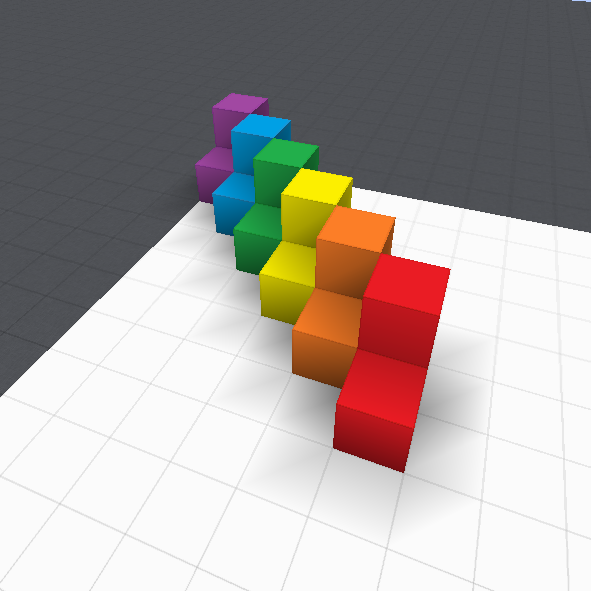}
    }\qquad 
    \subfigure{%
     \label{struct_d}
     \includegraphics[width=0.15\linewidth]{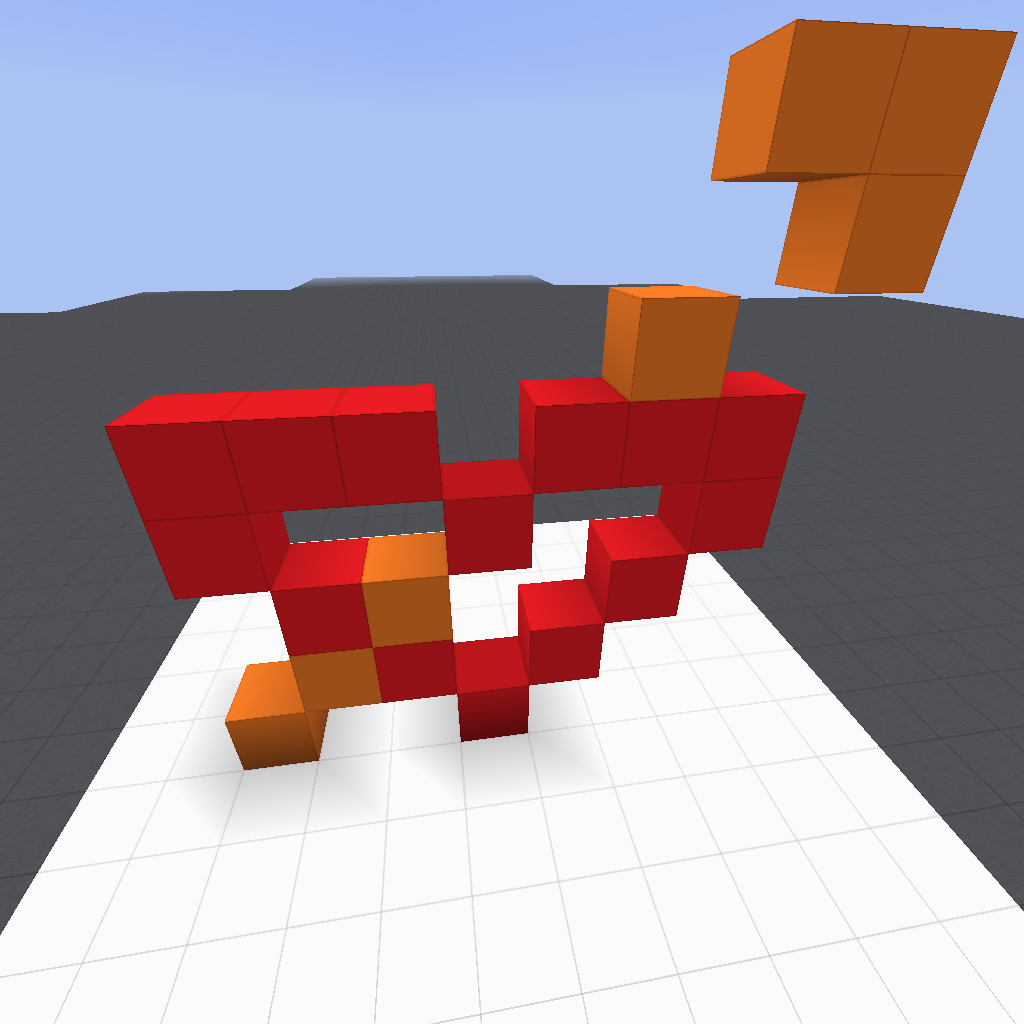}
    }
 }
\vspace{-0.5cm} 
\end{wrapfigure}


\paragraph{Architect simulator for training the silent Builder}
The correct instruction sequence to achieve the final goal is known since the target structures are associated with dialogs. We annotated each dialog's sub-goals and stored them in a queue, where each sub-goal corresponds to one specific step instruction from the Architect. We pop up a sub-task (e.g., in about the middle, build a column five tall) and wait until the agent completes it. If the agent completes this sub-task, we pop up the next sub-task. We trained a matching model to decide if the current sub-task is finished.

\subsubsection{Evaluation}
\label{sec:eval_rl}

For each subtask, the submitted agent is evaluated on the environment, with fixed initial and target states. We used the following three metrics: 
\begin{itemize}[leftmargin=*, nosep]
    \item The reward score $S_r$ is based on the average reward received by the agent in evaluation episodes, which is calculated as follows: 
    $ S_{r}=\frac{1}{N} \sum_{i=1}^{N}g_i$,
    where $N$ is a number of evaluation episodes, $g_i$ is episode reward, defined by $g_i=\sum_{t=1}^{T}r_t$.
    \item The success score $S_s$ indicates the number completely solved subtasks:
            $S_{s}=\frac{1}{N}\sum_{i=1}^{N}c_i$, 
        where 
        \begin{equation*}
        c_i = 
         \begin{cases}
           +1, &\text{if success,} \\
           0, &\text{otherwise.}
         \end{cases}
        \end{equation*}
    
    \item Completion rate score $S_{c}$:
            $S_{c}=\frac{1}{N}\sum_{i=1}^{N}1-\rho_i$, 
        where $\rho$ is a normalized Hamming distance between target and built structures.
\end{itemize}

\subsubsection{Baselines}
\label{sec:rl_baseline}
For the Silent Builder track, we provided several baselines, each training for a single target structure. First, we built grid-based agents implemented in RLlib \citep{rllib} framework. Agents were implemented with APE-X \citep{apex} and Rainbow \citep{rainbow_dqn} algorithms. Due to the combinatorial nature of the task, grid-baselines were able to solve only simplest possible tasks such as 5 block L-shaped structure. Second, we implemented a purely visual model-free agent using IMPALA algorithm \citep{impala} implemented with RLlib. The IMPALA agent yields a better performance achieving the normalized average reward of 12 in 20 million steps. This corresponds to 12 correctly placed blocks in the 18 blocks task. Last, we provided a visual model-based baseline agent which uses the Dreamer algorithm \citep{dreamer,dreamerv2}. The Dreamer agent shows similar performance to the model-free IMPALA agent but requires only 2 million steps to converge. The baselines are open sourced on Github\footnote{APE-X, Rainbow, IMPALA: \href{https://github.com/iglu-contest/iglu-builder-baseline-rllib}{https://github.com/iglu-contest/iglu-builder-baseline-rllib}}$^,$\footnote{Dreamer: \href{https://github.com/iglu-contest/iglu-builder-baseline-dreamer}{https://github.com/iglu-contest/iglu-builder-baseline-dreamer}}.

\subsubsection{Winning solutions}

\paragraph{Team 1: Hybrid Intelligence}

The Hybrid Intelligence team competed in the Builder track. Their solution improves upon the random agent by inducing a more accurate color distribution. They first extract a set of colors from the chat data via multilabel classification. When the random agent samples a hotbar action, which contains a color choice, with probability $1-\epsilon$ they instead equip a color sampled at random from the extracted set. This allows the agent to explore with probability $\epsilon$ via equipping colors that have not been extracted by the multilabel classification.

\paragraph{Team 2: NeuroAI}

The NeuroAI team trained a three stage compound model, which consists of: the pretrained 3D goal grid predictor that uses the text conversation as input, a convolutional autoencoder pretrained on random episodes, a policy that acts on a combination of the outputs of the previous two. The text-to-target predictor fine-tunes a head on top of the frozen Distil-RoBERTa model. The policy uses cross-attention with image embeddings as keys and values and goal grid as query followed by action and value heads. The agent was trained with PPO on an augmented set of tasks with a custom curriculum. 

\subsubsection{Results}
\paragraph{Silent Builder Track Results}By the end of the main competition, the Hybrid Intelligence team was placed first with a completion rate score of $0.365$. The NeuroAI team took second place with the leading score of $0.34$. In total, there were $96$ submissions and $37$ registered participants.

\paragraph{Human Evaluation}
After the main challenge, we ran human evaluations on the best agents submitted by both teams. We evaluated the agent's ability to respond to specific instructions correctly. To do this, each game episode was initialized with a prefix of dialog utterances and a partially built structure that corresponded to this dialog prefix. The last utterance defined an open goal with a result not presented in partial structure. With all that given, we measured the ability of the agent to respond to the instruction with actions correctly. 
Three judges performed the evaluation. We revealed a series of instructions to each of them. The instructions were combined with a pair of videos with action responses of each of the agents. Evaluators had to vote for the agent, which they thought performed better. Evaluators did not know which agent each animation corresponds to. The human evaluation confirmed the results of the automatic one, with the first team obtaining an average vote rate of 0.88 and Krippendorff's disagreement score of 0.56.

\section{Conclusions and Future Work}
IGLU has proposed unique and novel directions towards building collaborative embodied agents. In the course of the challenge, we have collected a new dataset suitable for NLP- and RL-related experiments. We developed a gym-like environment for training agents that takes instructions in the natural language as part of the observation, which should enable research in this area. The results of the completion clearly shows that the proposed direction is challenging and requires more exploration in the near future.

As for future direction, we will focus on investing more into the designing methodology for the builder automatic evaluation as well as human-in-the-loop one and specifying task for the interactive builder.
\section*{Acknowledgement}
We thank our sponsors Microsoft, AIRI, and Google. We would like to thank our advisory board: Julia Hockenmaier, Bill Dolan, Ryen W. White, Maarten de Rijke, and Sharada Mohanty.

\bibliography{ref}

\end{document}